\documentclass[11pt,a4paper,usenames,dvipsnames]{article}
\usepackage[hyperref]{acl2020}
\usepackage{times}
\usepackage{latexsym}

\usepackage{placeins}

\usepackage{microtype}

\aclfinalcopy 

\usepackage{graphicx}
\usepackage[T1]{fontenc}  

\newcommand\sbullet[1][.5]{\mathbin{\vcenter{\hbox{\scalebox{#1}{$\bullet$}}}}}

\usepackage{soul}

\title{A Large-Scale Multi-Document Summarization Dataset from the Wikipedia Current Events Portal}

\author{Demian Gholipour Ghalandari$^{1,2}$, Chris Hokamp$^1$, Nghia The Pham$^1$, \\
\textbf{John Glover}$^1$, \textbf{Georgiana Ifrim}$^2$\\
$^1$Aylien Ltd., Dublin, Ireland \\
$^2$Insight Centre for Data Analytics, University College Dublin, Ireland \\
$^1$\texttt{\{first-name\}@aylien.com} \\
\texttt{georgiana.ifrim@insight-centre.org}} 

\date{}

\begin{document}
\maketitle
\begin{abstract}
Multi-document summarization (MDS) aims to compress the content in large document collections into short summaries and has important applications in story clustering for newsfeeds, presentation of search results, and timeline generation.
However, there is a lack of datasets that realistically address such use cases at a scale large enough for training supervised models for this task.
This work presents a new dataset for MDS that is large both in the total number of document clusters and in the size of individual clusters.
We build this dataset by leveraging the Wikipedia Current Events Portal (WCEP), which provides concise and neutral human-written summaries of news events, with links to external source articles.
We also automatically extend these source articles by looking for related articles in the Common Crawl archive.
We provide a quantitative analysis of the dataset and empirical results for several state-of-the-art MDS techniques. 
The dataset is available at \url{https://github.com/complementizer/wcep-mds-dataset}. 
\end{abstract}

\section{Introduction}
Text summarization has recently received increased attention with the rise of deep learning-based end-to-end models, both for extractive and abstractive variants. However, so far, only single-document summarization has profited from this trend. Multi-document summarization (MDS) still suffers from a lack of established large-scale datasets.
This impedes the use of large deep learning models, which have greatly improved the state-of-the-art for various supervised NLP problems \cite{vaswani2017attention, paulus2018a, devlin2019bert}, and makes a robust evaluation difficult.
 Recently, several larger MDS datasets have been created: \citet{zopf2018auto, liu2018generating, multinews2019}. However, these datasets do not realistically resemble use cases with large automatically aggregated collections of news articles, focused on particular news events. This includes news event detection, news article search, and timeline generation. Given the prevalence of such applications, there is a pressing need for better datasets for these MDS use cases.  

\bgroup
\def\arraystretch{1.2}%
\begin{table}[t!]
\scriptsize    
\begin{tabular}{|p{0.95\columnwidth}|}
\hline
\textbf{Human-written summary}\\
\textcolor{RoyalPurple}{\textbf{Emperor Akihito abdicates the Chrysanthemum Throne in favor of his elder son, Crown Prince Naruhito. He is the first Emperor to abdicate in over two hundred years, since Emperor K\"{o}kaku in 1817.}}
\\
\hline 
\textbf{Headlines of source articles (WCEP)}\\
$\sbullet[.75]$ \textcolor{OliveGreen}{\textbf{Defining the Heisei Era: Just how peaceful were the past 30 years?}} \\
$\sbullet[.75]$ \textcolor{OliveGreen}{\textbf{As a New Emperor ls Enthroned in Japan, His Wife Won't Be Allowed to Watch}} \\ \hline

\textbf{Sample Headlines from Common Crawl}\\
$\sbullet[.75]$ \textcolor{RedViolet}{\textbf{Japanese Emperor Akihito to abdicate after three decades on throne}} \\
$\sbullet[.75]$ \textcolor{RedViolet}{\textbf{Japan's Emperor Akihito says he is abdicating as of Tuesday at a ceremony, in his final official address to his people}} \\
$\sbullet[.75]$ \textcolor{RedViolet}{\textbf{Akihito begins abdication rituals as Japan marks end of era}} \\
\hline
\end{tabular}
\caption{Example event summary and linked source articles from the Wikipedia Current Events Portal, and additional extracted articles from Common Crawl.}
\label{table:intro-example}
\end{table}
\egroup

In this paper, we present the Wikipedia Current Events Portal (WCEP) dataset, which is designed to address real-world MDS use cases. The dataset consists of 10,200 clusters with one human-written summary and 235 articles per cluster on average. 
We extract this dataset starting from the Wikipedia Current Events Portal (WCEP)\footnote{\url{https://en.wikipedia.org/wiki/Portal:Current_events}}. Editors on WCEP write short summaries about news events and provide a small number of links to relevant source articles. 
We extract the summaries and source articles from WCEP and increase the number of source articles per summary by searching for similar articles in the Common Crawl News dataset\footnote{\url{https://commoncrawl.org/2016/10/news-dataset-available/}}. As a result, we obtain large clusters of highly redundant news articles, resembling the output of news clustering applications. Table \ref{table:intro-example} shows an example of an event summary, with headlines from both the original article and from a sample of the associated additional sources. 
In our experiments, we test a range of unsupervised and supervised MDS methods to establish baseline results. We show that the additional articles lead to much higher upper bounds of performance for standard extractive summarization, and help to increase the performance of baseline MDS methods.

We summarize our contributions as follows: 
\begin{itemize}
    \itemsep0em 
    \item We present a new large-scale dataset for MDS, that is better aligned with several real-world industrial use cases.
    \item We provide an extensive analysis of the properties of this dataset. 
    \item We provide empirical results for several baselines and state-of-the-art MDS methods aiming to facilitate future work on this dataset.
\end{itemize}

\section{Related Work}

\subsection{Multi-Document Summarization}
Extractive MDS models commonly focus on either ranking sentences by importance \cite{hong2014improving, cao2015ranking, yasunaga2017graph} or on global optimization to find good combinations of sentences, using heuristic functions of summary quality \cite{gillick2009scalable, lin2011class, peyrard2016general}.

Several abstractive approaches for MDS are based on multi-sentence compression and sentence fusion \cite{ganesan2010opinosis, banerjee2015multi, chali2017towards, nayeem2018abstractive}. Recently, neural sequence-to-sequence models, which are the state-of-the-art for abstractive single-document summarization \cite{rush2015neural, nallapati2016abstractive, see2017get},  have been used for MDS, e.g., by applying them to extractive summaries \cite{liu2018generating} or by directly encoding multiple documents \cite{zhang2018towards, multinews2019}. 

\subsection{Datasets for MDS}
Datasets for MDS consist of clusters of source documents and at least one ground-truth summary assigned to each cluster. Commonly used traditional datasets include the \textbf{DUC 2004}  \cite{paul2004introduction} and \textbf{TAC 2011} \cite{owczarzak2011overview}, which consist of only 50 and 100 document clusters with 10 news articles on average. The \textbf{MultiNews} dataset \cite{multinews2019} is a recent large-scale MDS dataset, containing 56,000 clusters, but each cluster contains only 2.3 source documents on average. The sources were hand-picked by editors and do not reflect use cases with large automatically aggregated document collections. MultiNews has much more verbose summaries than WCEP.

\citet{zopf2018auto} created the \textbf{auto-\textit{h}MDS} dataset by using the lead section of Wikipedia articles as summaries, and automatically searching for related documents on the web, resulting in 7,300 clusters. The \textbf{WikiSum} dataset \cite{liu2018generating} uses a similar approach and additionally uses cited sources on Wikipedia. The dataset contains 2.3 million clusters.
These Wikipedia-based datasets also have long summaries about various topics, whereas our dataset focuses on short summaries about news events.

\section{Dataset Construction}
\paragraph{Wikipedia Current Events Portal:}
WCEP lists current news events on a daily basis. Each news event is presented as a summary with at least one link to external news articles. According to the editing guidelines\footnote{\url{https://en.wikipedia.org/wiki/Wikipedia:How_the_Current_events_page_works}}, the summaries must be short, up to 30-40 words, and written in complete sentences in the present tense, avoiding opinions and sensationalism. Each event must be of international interest. Summaries are written in English, and news sources are preferably English. 

\paragraph{Obtaining Articles Linked on WCEP:}
We parse the WCEP monthly pages to obtain a list of individual events, each with a list of URLs to external source articles.
To prevent the source articles of the dataset from becoming unavailable over time, we use the `Save Page Now` feature of the Internet Archive\footnote{\url{https://web.archive.org/save/}}. We request snapshots of all source articles that are not captured in the Internet Archive yet. 
We download and extract all articles from the Internet Archive Wayback Machine\footnote{\url{https://archive.org/web/}} using the newspaper3k\footnote{\url{https://github.com/codelucas/newspaper}} library. 

\paragraph{Additional Source Articles:} Each event from WCEP contains only 1.2 sources on average, meaning that most editors provide only one source article when they add a new event. In order to extend the set of input articles for each of the ground-truth summaries, we search for similar articles in the Common Crawl News dataset\footnote{\url{https://commoncrawl.org/2016/10/news-dataset-available/}}. 

We train a logistic regression classifier to decide whether to assign an article to a summary, using the original WCEP summaries and source articles as training data. 
For each event, we label the \texttt{article-summary} pair for each source article of the event as positive. We create negative examples by pairing each event with source articles from other events of the same date, resulting in a positive-negative ratio of 7:100. The features used by the classifier are listed in Table \ref{tab:features}.
\begin{table}[h!]
\small
\resizebox{0.9\columnwidth}{!}{%
\begin{tabular}{|l|}
\hline
\texttt{tf-idf} similarity between title and summary \\ \hline
\texttt{tf-idf} similarity between body and summary \\ \hline
No. entities from summary appearing in title \\ \hline
No. linked entities from summary appearing in body \\ \hline
\end{tabular}%
}
\caption{Features used in the \texttt{article-summary} binary classifier.}
\label{tab:features}
\end{table}

We use unigram bag-of-words vectors with TF-IDF weighting and cosine similarity for the first two features. The entities are phrases in the WCEP summaries that the editors annotated with hyperlinks to other Wikipedia articles. We search for these entities in article titles and bodies by exact string matching. The classifier achieves $90\%$ Precision and $74\%$ Recall of positive examples on a hold-out set.

For each event in the original dataset, we apply the classifier to articles published in a window of $\pm 1$ days of the event date and add those articles that pass a classification probability of 0.9. If an article is assigned to multiple events, we only add it to the event with the highest probability. This procedure increases the number of source articles per summary considerably (Table \ref{tab:cluster-stats}).

\paragraph{Final Dataset:} Each example in the dataset consists of a ground-truth summary and a cluster of original source articles from WCEP, combined with additional articles from Common Crawl. The dataset has 10,200 clusters, which we split roughly into 80\% training, 10\% validation and 10\% test (Table \ref{tab:overview-stats}). The split is done chronologically, such that no event dates overlap between the splits. We also create a truncated version of the dataset with a maximum of 100 articles per cluster, by retaining all original articles and randomly sampling from the additional articles. 

\section{Dataset Statistics and Analysis}

\subsection{Overview}

Table \ref{tab:overview-stats} shows the number of clusters and of articles from all clusters combined, for each dataset partition. Table \ref{tab:cluster-stats} shows statistics for individual clusters. We show statistics for the entire dataset (WCEP-total), and for the truncated version (WCEP-100) used in our experiments. The high mean cluster size is mostly due to articles from Common Crawl.

\begin{table}[htb!]
\resizebox{\columnwidth}{!}{%
\begin{tabular}{|l|l|l|l|l|}
\hline
 & \textsc{train} & \textsc{val} & \textsc{test} & \textsc{total} \\ \hline
\# clusters & 8,158 & 1,020 & 1,022 & 10,200 \\ \hline
\# articles (WCEP-total) & 1.67m & 339k & 373k & 2.39m  \\ \hline
\# articles (WCEP-100) & \~494k & 78k & 78k & 650k  \\ \hline
period start & 2016-8-25 & 2019-1-6 & 2019-5-8 & - \\ \hline
period end & 2019-1-5 & 2019-5-7 & 2019-8-20 & - \\ \hline
\end{tabular}}
\caption{Size overview of the WCEP dataset.}
\label{tab:overview-stats}
\end{table}

\begin{table}[htb!]
\small
\resizebox{\columnwidth}{!}{%
\begin{tabular}{|l|l|l|l|l|}
\hline
            & \textsc{min} & \textsc{max} & \textsc{mean} & \textsc{median}\\ \hline
\# articles (WCEP-total) & 1 & 8411 & 234.5 & 78 \\ \hline            
\# articles (WCEP-100) & 1 & 100 & 63.7 & 78  \\ \hline
\# WCEP articles & 1 & 5 & 1.2 & 1  \\ \hline
\# summary words & 4 & 141 & 32 & 29  \\ \hline
\# summary sents & 1 & 7 & 1.4 & 1 \\ \hline
\end{tabular}}
\caption{Stats for individual clusters in WCEP dataset.}
\label{tab:cluster-stats}
\end{table}

\subsection{Quality of Additional Articles}
To investigate how related the additional articles obtained from Common Crawl are to the summary they are assigned to, we randomly select 350 for manual annotation. We compare the  article title and the first three sentences to the assigned summary, and pick one of the following three options: 1) "on-topic" if the article focuses on the event described in the summary, 2) "related" if the article mentions the event, but focuses on something else, e.g., follow-up, and 3) "unrelated" if there is no mention of the event. This results in 52\% on-topic, 30\% related, and 18\% unrelated articles. We think that this amount of noise is acceptable, as it resembles noise present in applications with automatic content aggregation. Furthermore, summarization performance benefits from the additional articles in our experiments (see Section \ref{sec:experiments}).

\subsection{Extractive Strategies}
\label{sec:extractive strategies}

Human-written summaries can vary in the degree of how extractive or abstractive they are, i.e., how much they copy or rephrase information in source documents. To quantify \textit{extractiveness} in our dataset, we use the measures \textit{coverage} and \textit{density}  defined by \citet{grusky2018newsroom}:
\begin{equation}
\small
Coverage(A, S) = \frac{1}{|S|} \sum_{f \in F(A, S)} |f|
\end{equation}
\begin{equation}
\small
    Density(A, S) = \frac{1}{|S|} \sum_{f \in F(A, S)} |f|^2
\end{equation}

Given an article $A$ consisting of tokens $\langle a_1, a_2, ..., a_n \rangle$ and its summary $S=\langle s_1, s_2, ..., s_n \rangle$, $F(A, S)$ is the set of token sequences (fragments) shared between $A$ and $S$, identified in a greedy manner. Coverage measures the proportion of words from the summary appearing in these fragments. Density is related to the average length of shared fragments and measures how well a summary can be described as a series of extractions. In our case, $A$ is the concatenation of all articles in a cluster. 

\begin{figure}[htb!]
  \centering
  \includegraphics[width=1\columnwidth]{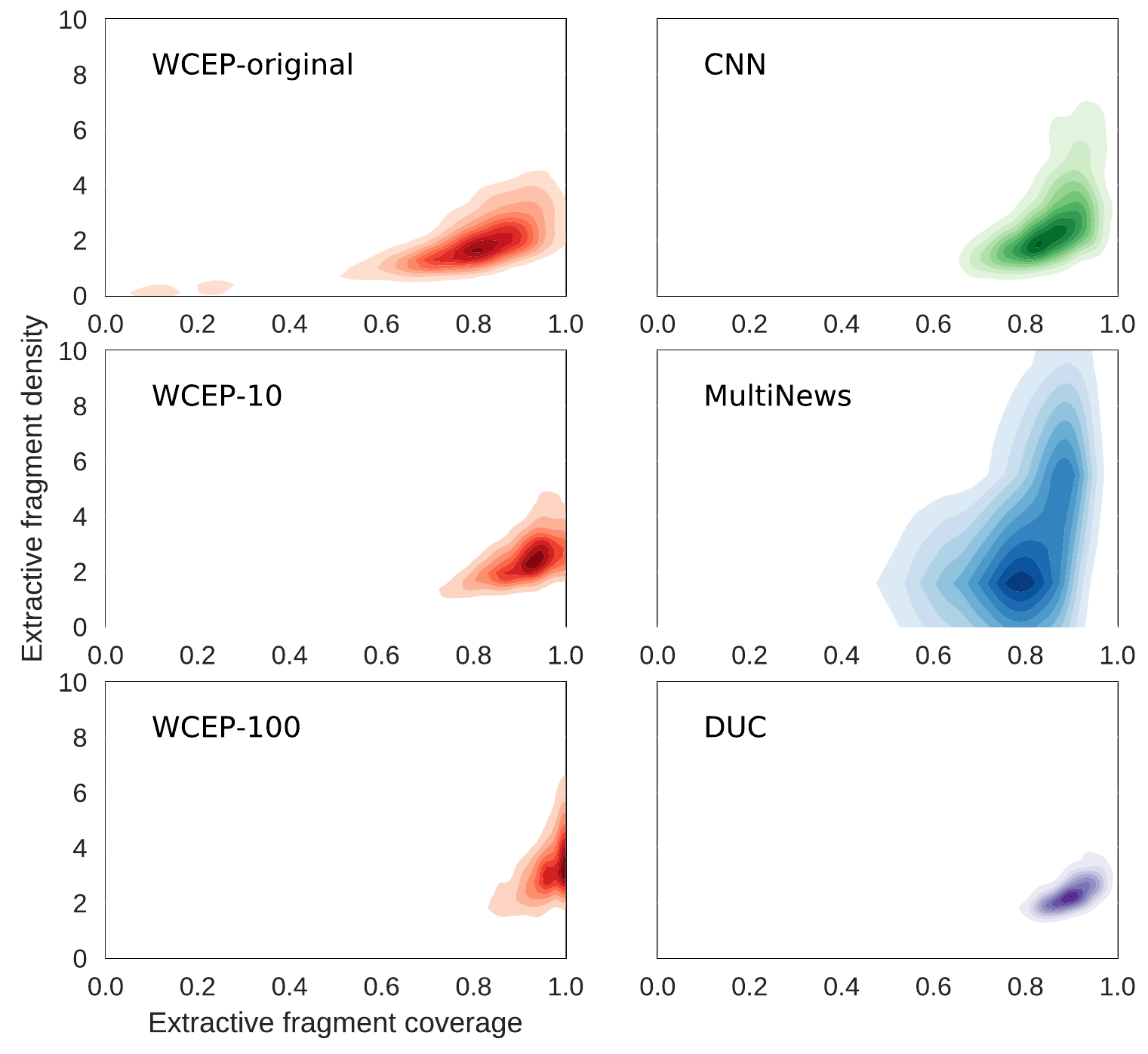}
  \caption{Coverage and density on different summarization datasets.}
  \label{fig:kde_plots}
\end{figure}
Figure \ref{fig:kde_plots} shows the distribution of coverage and density in different summarization datasets. WCEP-10 refers to a truncated version of our dataset with a maximum cluster size of 10. The WCEP dataset shows increased coverage if more articles from Common Crawl are added, i.e., all words of a summary tend to be present in larger clusters. High coverage suggests that retrieval and copy mechanisms within a cluster can be useful to generate summaries. Likely due to the short summary style and editor guidelines, high density, i.e., copying of long sequences, is not as common in WCEP as in the MultiNews dataset. 

\section{Experiments}
\label{sec:experiments}

\subsection{Setup}
Due to scalability issues of some of the tested methods, we use the truncated version of the dataset with a maximum of 100 articles per cluster (WCEP-100). The performance of the methods that we consider starts to plateau after 100 articles (see Figure \ref{fig:sizes}). We set a maximum summary length of 40 tokens, which is in accordance with the editor guidelines in WCEP. This limit also corresponds to the optimal length of an extractive oracle optimizing ROUGE F1-scores\footnote{We tested lengths 25 to 50 in steps of 5. For these tests, the oracle is forced to pick a summary up to that length.}. We recommend to evaluate models with a dynamic (potentially longer) output length using F1-scores and optionally to provide Recall results with truncated summaries.
Extractive methods should only return lists of full untruncated sentences up to that limit. We evaluate lowercased versions of summaries and do not modify ground-truth or system summaries otherwise. We compare and evaluate systems using F1-score and Recall of ROUGE-1, ROUGE-2, and ROUGE-L \cite{lin-2004-rouge}. In the following, we abbreviate ROUGE-1 F1-score and Recall with R1-F and R1-R, etc.
\subsection{Methods}
We evaluate the following oracles and baselines to put evaluation scores into perspective:
\begin{itemize}
    \itemsep0em 
    \item \textsc{Oracle (Multi)}: Greedy oracle, adds sentences from a cluster that optimize R1-F of the constructed summary until R1-F  decreases.
    \item \textsc{Oracle (Single)}: Best of oracle summaries extracted from individual articles in a cluster.
    \item \textsc{Lead Oracle}: The lead (first sentences up to 40 words) of an individual article with the best R1-F score within a cluster.
    \item \textsc{Random Lead}: The lead of a randomly selected article, which is our alternative to the lead baseline used in single-document summarization.
\end{itemize} 
We evaluate the unsupervised methods \textsc{TextRank} \cite{mihalcea2004textrank}, \textsc{Centroid} \cite{radev2004centroid} and \textsc{Submodular} \cite{lin2011class}. We test the following supervised methods:
\begin{itemize}
    \itemsep0em 
    \item \textsc{TSR}: Regression-based sentence ranking using statistical features and averaged word embeddings \cite{ren-etal-2016-redundancy}.
    \item \textsc{BertReg}: Similar framework to \textsc{TSR} but with sentence embeddings computed by a pre-trained BERT model \cite{devlin2019bert}. Refer to Appendix \ref{appendix-bertreg} for more details.
\end{itemize}
We tune hyperparameters of the methods described above on the validation set of WCEP-100 (Appendix \ref{appendix-ext-details}). We also test a simple abstractive baseline, \textsc{Submodular + Abs}: We first create an extractive multi-document summary with a maximum of 100 words using \textsc{Submodular}. We pass this summary as a pseudo-article to the abstractive \textit{bottom-up attention} model \cite{gehrmann2018bottom} to generate the final summary. We use an implementation from OpenNMT\footnote{ \url{https://opennmt.net/OpenNMT-py/Summarization.html}}
with a model pre-trained on the CNN/Daily Mail dataset. All tested methods apart from \textsc{Oracle (Multi \& Single)} observe the length limit of 40 tokens.

\subsection{Results}
Table \ref{tab:evaluation} presents the results on the WCEP test set. The supervised methods \textsc{TSR} and \textsc{BertReg} show advantages over unsupervised methods, but not by a large margin, which poses an interesting challenge for future work. The high extractive bounds defined by \textsc{Oracle (Single)} suggest that identifying important documents before summarization can be useful in this dataset. The dataset does not favor lead summaries: \textsc{Random Lead} is of low quality, and \textsc{Lead Oracle} has relatively low F-scores (although very high Recall). The \textsc{Submodular + Abs} heuristic for applying a pre-trained abstractive model does not perform well. 

\begin{table}[]
\centering
\small
\begin{tabular}{|l|lll|}
\hline
 \multicolumn{4}{|c|}{\textbf{F-score}} \\ \hline
 Method & R1 & R2 & RL \\ \hline
\textsc{Oracle (Multi)} & 0.558 & 0.29 & 0.4  \\
\textsc{Oracle (Single)} & 0.539 & 0.283 & 0.401  \\
\textsc{Lead Oracle} & 0.329 & 0.131 & 0.233  \\
\textsc{Random Lead} & 0.276 & 0.091 & 0.206  \\ 
\textsc{Random} & 0.181 & 0.03 & 0.128  \\ \hline
\textsc{TextRank} & 0.341 & 0.131 & 0.25  \\
\textsc{Centroid} & 0.341 & 0.133 & 0.251  \\
\textsc{Submodular} & 0.344 & 0.131 & 0.25  \\
\textsc{TSR} & \textbf{0.353} & \textbf{0.137} & \textbf{0.257}  \\
\textsc{BertReg} & 0.35 & 0.135 & 0.255 \\
\textsc{Submodular+Abs} & 0.306 & 0.101 & 0.214  \\ \hline
 \multicolumn{4}{|c|}{\textbf{Recall}} \\ \hline
Method & R1 & R2 & RL \\ \hline
\textsc{Oracle (Multi)} & 0.645 & 0.331 & 0.458  \\
\textsc{Oracle (Single)} & 0.58 & 0.304 & 0.431  \\
\textsc{Lead Oracle} & 0.525 & 0.217 & 0.372  \\
\textsc{Random Lead} & 0.281 & 0.094 & 0.211  \\
\textsc{Random} & 0.203 & 0.034 & 0.145 \\ \hline
\textsc{TextRank} & 0.387 & 0.152 & 0.287  \\
\textsc{Centroid} & 0.388 & 0.154 & 0.29  \\
\textsc{Submodular} & 0.393 & 0.15 & 0.289  \\
\textsc{TSR} & \textbf{0.408} & \textbf{0.161} & \textbf{0.301}  \\
\textsc{BertReg} & 0.407 & 0.16 & \textbf{0.301}  \\
\textsc{Submodular+Abs} & 0.363 & 0.123 & 0.258  \\ \hline
\end{tabular}
\caption{Evaluation results on test set.}
\label{tab:evaluation}
\end{table}

\subsection{Effect of Additional Articles}
\begin{figure}[htb!]
  \centering
  \includegraphics[width=1\columnwidth]{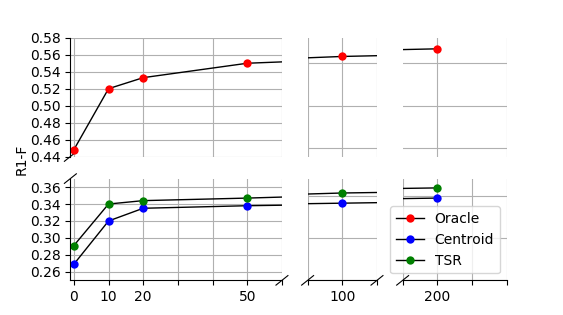}
  \caption{ROUGE-1 F1-scores for different numbers of supplementary articles from Common Crawl.}
  \label{fig:sizes}
\end{figure}

Figure \ref{fig:sizes} shows how the performance of several methods on the test set increases with different amounts of additional articles from Common Crawl. Using 10 additional articles causes a steep improvement compared to only using the original source articles from WCEP. However, using more than 100 articles only leads to minimal gains.

\section{Conclusion}

We present a new large-scale MDS dataset for the news domain, consisting of large clusters of news articles, associated with short summaries about news events. We hope this dataset will facilitate the creation of real-world MDS systems for use cases such as summarizing news clusters or search results. We conducted extensive experiments to establish baseline results, and we hope that future work on MDS will use this dataset as a benchmark. Important challenges for future work include how to scale deep learning methods to such large amounts of source documents and how to close the gap to the oracle methods. 

\section*{Acknowledgments}
This work was funded by the Irish Research Council (IRC) under grant number EBPPG/2018/23, the Science Foundation Ireland (SFI) under grant number 12/RC/2289\_P2 and the enterprise partner Aylien Ltd. 

\bibliographystyle{ACM-Reference-Format}
\bibliography{references}

\appendix
\section{Appendices}

\subsection{\textsc{BertReg}}
\label{appendix-bertreg}
This method uses a regression model to score and rank sentences. For a particular sentence, we obtain a contextualized embedding from a pre-trained BERT model\footnote{We use the 12-layer model from  \url{https://github.
com/hanxiao/bert-as-service}}.  We concatenate the embedding with several
statistical and surface-form sentence features shown in Table \ref{tab:bertreg-features}.

\begin{table}[!h]
\small
\centering
\begin{tabular}{|l|} \hline
length (in tokens) \\ \hline
position \\ \hline
stop word ratio \\ \hline
mean \texttt{tf} \\ \hline
mean \texttt{tf-idf} \\ \hline
mean \texttt{tf-icf} \\ \hline
mean \texttt{cluster-df} \\ \hline
\end{tabular}
\caption{Features used for \textsc{BertReg} apart from the contextual sentence embeddings.}
\label{tab:bertreg-features}
\end{table}

The corpus-level document and cluster frequencies (cf) in \texttt{tf-idf} and \texttt{tf-icf} are obtained
from the training set. \texttt{cluster-df} refers to the document frequency within a particular cluster. We feed this concatenated sentence vector to a feed-forward network with one hidden layer of size 256. The model is trained to predict the R1 F-score
between a sentence and the summary of a cluster,
using the mean squared error loss. We found the
F-score to work better than Precision or Recall. We
use the SGD optimizer, a learning rate of 0.02, and
train for 8 epochs with batch size 8. To construct a
summary, we predict scores using this model, rank
sentences, and greedily pick sentences from the
ranked list under a redundancy constraint, as used
in \textsc{TSR}.

\subsection{Implementation Details for Extractive Methods}
\label{appendix-ext-details}
We implement the methods \textsc{TextRank}, \textsc{Centroid}, \textsc{TSR} and \textsc{BertReg} in a commonly used framework that greedily selects sentences from a ranked list while avoiding redundancy \cite{zopf2018scores}. We measure \texttt{redundancy} as the proportion of bigrams in a new sentence that appear in an already selected sentence. For each method, we tune threshold values for \texttt{redundancy} from 0 to 1 in steps of 0.1. For \textsc{Submodular}, we tune a parameter called \texttt{diversity} with values 1 to 10 in steps of 1, which has a similar role as the \texttt{redundancy} threshold. We use 100 randomly selected clusters from the validation set in WCEP-100 for parameter tuning. We set a minimum sentence length of 7 tokens which avoids summaries slighly shorter than the 40 token limit to be padded with very short or broken sentences. 
\end{document}